\pdfoutput=1

\documentclass[11pt]{article}

\usepackage[final]{acl}

\usepackage{times}
\usepackage{latexsym}

\usepackage[T1]{fontenc}

\usepackage[utf8]{inputenc}

\usepackage{microtype}

\usepackage{inconsolata}

\usepackage{graphicx}

\usepackage{diagbox} 
\usepackage{multirow}  
\usepackage{amssymb}  
\usepackage{adjustbox}  
\usepackage{amsmath}  
\usepackage{booktabs}  
\usepackage{threeparttable}  

\title{Pcc-tuning: Breaking the Contrastive Learning Ceiling in Semantic Textual Similarity}

\author{Bowen Zhang \and Chunping Li \\
  School of Software, Tsinghua University \\
  \texttt{zbw23@mails.tsinghua.edu.cn}, \texttt{cli@tsinghua.edu.cn} \\}

\begin{document}
\maketitle
\begin{abstract}
Semantic Textual Similarity (STS) constitutes a critical research direction in computational linguistics and serves as a key indicator of the encoding capabilities of embedding models. Driven by advances in pre-trained language models and contrastive learning, leading sentence representation methods have reached an average Spearman's correlation score of approximately 86 across seven STS benchmarks in SentEval. However, further progress has become increasingly marginal, with no existing method attaining an average score higher than 86.5 on these tasks. This paper conducts an in-depth analysis of this phenomenon and concludes that the upper limit for Spearman's correlation scores under contrastive learning is 87.5. To transcend this ceiling, we propose an innovative approach termed Pcc-tuning, which employs \textbf{\underline{P}}earson's \textbf{\underline{c}}orrelation \textbf{\underline{c}}oefficient as a loss function to refine model performance beyond contrastive learning. Experimental results demonstrate that Pcc-tuning can markedly surpass previous state-of-the-art strategies with only a minimal amount of fine-grained annotated samples. \footnote{Our code and checkpoints are available at \url{https://github.com/ZBWpro/Pcc-tuning}.}
\end{abstract}

\section{Introduction}

As a fundamental task within Natural Language Processing (NLP), Semantic Textual Similarity (STS) is not only widely applied across various real-world scenarios including text clustering, information retrieval, and dialogue systems, but also serves as a principal means for evaluating sentence embeddings \citep{SimCSE-EMNLP-2021}.

Sentence embeddings refer to vector encodings that encapsulate the semantic essence of original texts. Owing to their capacity to facilitate offline computation as well as their pivotal role in realizing retrieval-augmented generation \citep{RAG-survey-2024}, research in this area has garnered considerable attention from numerous institutions and scholars in recent years.

The quality of sentence embeddings is typically assessed via the SentEval \citep{SentEval-LREC-2018} toolkit, which measures models based on their average Spearman correlation across seven STS benchmarks. With the continuous advancement of pre-trained language models (PLMs), contrastive learning, and prompt engineering, cutting-edge works in this field have progressively elevated leaderboard scores from an initial 60 \citep{GloVe-EMNLP-2014} to around 86 \citep{PromptEOL-2023}. As a result, the "PLM + contrastive learning" framework has become the mainstream paradigm in sentence representation research.

However, as illustrated in Table~\ref{tab:sts_bound}, models' performance on standard STS tasks in SentEval appears to have hit a significant bottleneck. Whether utilizing classical discriminative PLMs like BERT \citep{BERT-NAACL-2019} or emerging generative PLMs such as LLaMA2 \citep{LLaMA2-2023} and Mistral \citep{Mistral-2023}, contemporary state-of-the-art (SOTA) strategies are unable to achieve Spearman's correlation scores exceeding 86.5. Moreover, despite variations in training datasets, contrastive learning objectives, and model architectures, the final performance are generally similar if the same type of PLM is selected.
\begin{table}[htbp]
\centering
\begin{tabular}{ccc}
    \toprule
    \bf Methods & \bf PLMs & \bf Spearman\\
    \midrule
    SimCSE & BERT$_{\rm 110m}$ & 81.57 \\ 
    PromptBERT & BERT$_{\rm 110m}$ & 81.97 \\ 
    PromCSE & BERT$_{\rm 110m}$ & 82.13 \\ 
    SuCLSE & BERT$_{\rm 110m}$ & 82.17 \\ 
    \midrule
    SimCSE $^\diamondsuit$ & LLaMA2$_{\rm 7b}$ & 85.24 \\ 
    PromptEOL $^\spadesuit$ & LLaMA2$_{\rm 7b}$ & 85.40 \\ 
    PromptSTH $^\spadesuit$ & LLaMA2$_{\rm 7b}$ & 85.41 \\ 
    PromptSUM $^\spadesuit$ & LLaMA2$_{\rm 7b}$ & 85.53 \\ 
    PromCSE $^\diamondsuit$ & LLaMA2$_{\rm 7b}$ & 85.70 \\ 
    AngIE $^\diamondsuit$ & LLaMA2$_{\rm 7b}$ & 85.96 \\ 
    DeeLM $^\diamondsuit$ & LLaMA2$_{\rm 7b}$ & 86.01 \\ 
    \midrule
    PromptEOL $^\spadesuit$ & Mistral$_{\rm 7b}$ & 85.50 \\ 
    PromptSTH $^\spadesuit$ & Mistral$_{\rm 7b}$ & 85.66 \\ 
    PromptSUM $^\spadesuit$ & Mistral$_{\rm 7b}$ & 85.83 \\ 
    \bottomrule
\end{tabular}
\caption{Average Spearman's correlation scores obtained by leading methods on the seven STS benchmarks collected in SentEval. $\diamondsuit$: results from \citep{DeeLM-2023}. $\spadesuit$: results from \citep{PretCoTandKE-ICIC-2024}.}
\label{tab:sts_bound}
\end{table}

In this regard, DeeLM \citep{DeeLM-2023} posits that PLMs may have reached their performance limits on STS tasks. However, this paper will demonstrate through rigorous mathematical derivation that the core factor causing this performance ceiling is not the inadequacy of PLMs, but inherent flaws in contrastive learning loss functions. Specifically, contrastive learning only distinguishes between two categories: similar and dissimilar, in determining the semantic relationships between text pairs. This binary classification strategy restricts its maximum achievable Spearman's correlation score to 87.5, even under optimal conditions.

Following this proof, we introduce Pcc-tuning, a novel approach that employs a two-stage training process. This method enhances models' semantic discrimination capabilities by leveraging a small amount of fine-grained annotated data post contrastive learning. With the same 7B-scale generative PLMs, Pcc-tuning can substantially surpass previous best results on the seven aforementioned STS tasks and break through the performance ceiling of 87.5.

The main contributions of this study are outlined as follows:
\begin{itemize}
\item By analyzing the theoretical limits of binary classifiers in STS tasks, we prove that the upper bound for Spearman's correlation scores using contrastive learning methods is 87.5. This finding effectively explains the performance plateau encountered by prior sentence representation research.

\item Building upon this, we propose Pcc-tuning, a method capable of taking full advantage of fine-grained labeled data with Pearson correlation as its loss function. After fine-tuning PLMs through contrastive learning, we only need to introduce annotated text pairs amounting to 1.96\% of the original training set to bring notable performance improvements.

\item We extensively validate the effectiveness of Pcc-tuning across internationally recognized STS benchmarks and multiple transfer tasks. Experimental results show that Pcc-tuning consistently outperforms existing SOTA methods across different PLMs, prompts and hyperparameter settings.
\end{itemize}

\section{Understanding the Performance Upper Bound of Contrastive Learning}
\label{sec:proof}

\subsection{Contrastive Learning and Binary Classifiers}
\label{sec:binary}

Currently, leading approaches for sentence representation predominantly center around contrastive learning, with InfoNCE Loss \citep{InfoNCE-2018} being the most commonly adopted loss function. Given an input text $x_i$, InfoNCE Loss computes the similarity between this sample and its positive example $x_i^+$ in the numerator, contrasting it with the similarity calculations between $x_i$ and other texts within the same batch in the denominator. This formulation aims to bring similar instances closer while pushing dissimilar ones apart. The mathematical expression for InfoNCE Loss is presented in Equation~\ref{eq:info_nce}, where $f(\cdot)$ denotes the encoding method, $N$ represents the batch size, and $\tau$ signifies a temperature hyperparameter.
\begin{equation}
     \mathcal{\ell}_{i} = - {\log} \frac{e^{{\cos}(f(x_i),f(x_i^+))/\tau}}{\sum_{j=1}^Ne^{{\cos}(f(x_i),f(x_j^+))/\tau}}
\label{eq:info_nce}
\end{equation}

Equation~\ref{eq:info_nce} indicates that contrastive learning loss functions, exemplified by InfoNCE Loss, essentially classify sentence pairs into two distinct classes: similar and dissimilar. However, no further distinctions are made within these two categories. In other words, as long as $x_i$ is semantically different from $x_j$ or $x_k$, InfoNCE Loss treats both $(x_i, x_j)$ and $(x_i, x_k)$ as negative sample pairs. As for which of $(x_i, x_j)$ and $(x_i, x_k)$ exhibits a lower degree of similarity, contrastive learning neither concerns itself with this information nor can it readily leverage such details. Indeed, for the majority of embedding models, their training sets are specially adjusted to provide coarse-grained categorical annotations, so as to better align with the contrastive learning framework \citep{SimCSE-EMNLP-2021, PromptBERT-EMNLP-2022, PromptEOL-2023}.

\begin{figure}[htbp]
\centering
\includegraphics[width=1.0\linewidth]{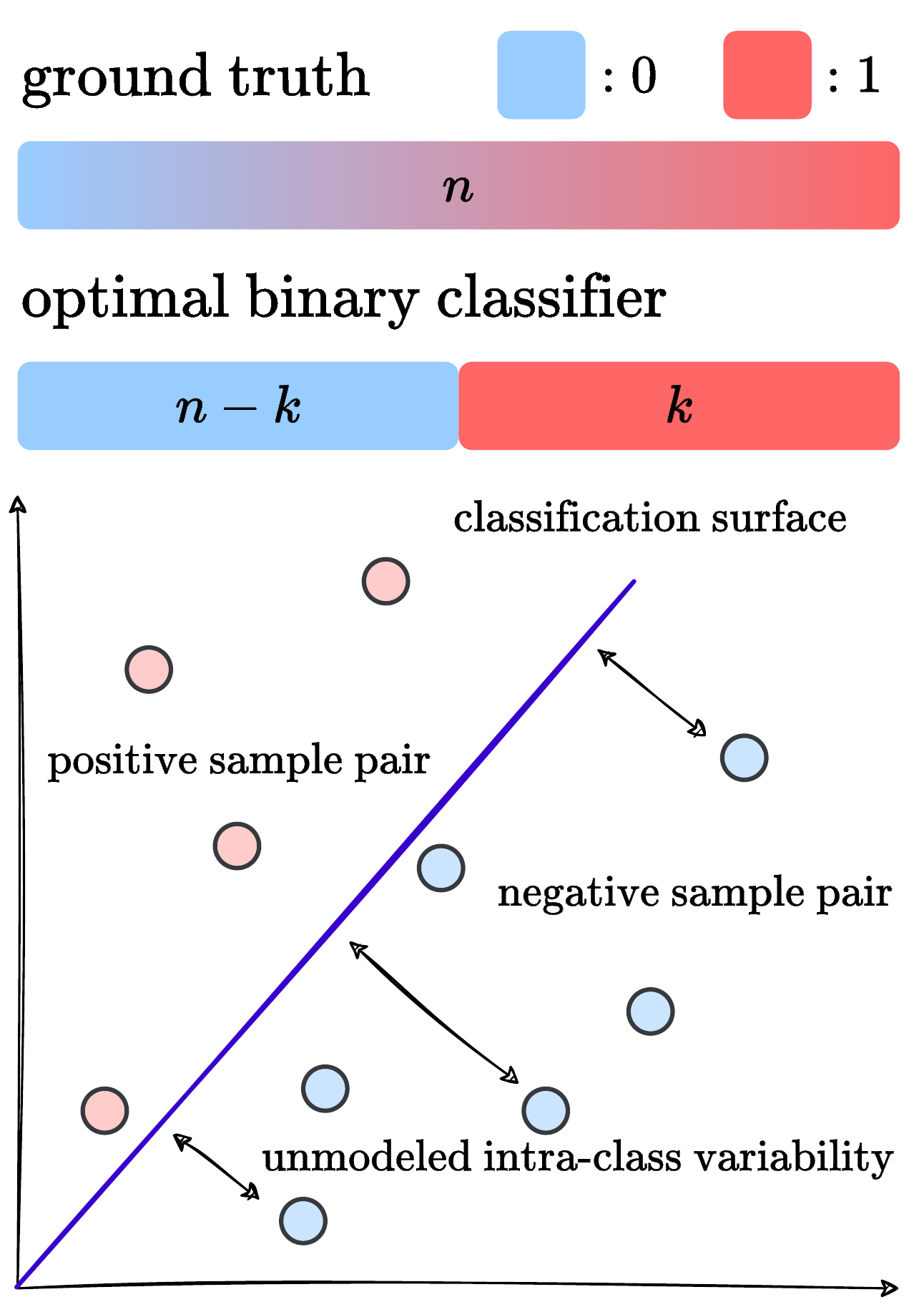}
\caption{Illustration of the operation of an optimal binary classifier in handling STS tasks. Although the actual similarity scores of the text pairs are a series of floating-point numbers, the binary classifier focuses solely on categorizing them into two classes: similar and dissimilar, without modeling the variability within each category.}
\label{fig:binary}
\end{figure}

Therefore, for a set of text pairs $\{(x_i, x_i^?)\}_1^n$, the optimal scenario for contrastive learning methods is to classify the $k$ most similar pairs as positive and the remaining $n-k$ pairs as negative. This setup ensures that there are no inversions in the predicted scores provided by the model. Such an ideal state for contrastive learning models functions similarly to an optimal binary classifier, as illustrated in Figure~\ref{fig:binary}. This classifier segments the dataset into two groups based on a threshold $k$, assigning a positive label to all samples above the threshold and a negative label to those below. Analyzing the efficacy of this binary classifier reveals the performance boundary of contrastive learning.

One potential concern with this analogy is that contrastive learning models compute cosine similarities between embeddings during the testing phase, resulting in continuous predicted values. Nevertheless, since the model does not differentiate between internal discrepancies within the positive and negative classes during training, it cannot be expected to possess the capability to discern fine-grained semantic similarities. As training data continues to flow in, InfoNCE Loss gradually guides the model toward the characteristics of an ideal binary classifier. However, constrained by the expressive power of neural networks, as well as the scale and quality of the training data, relying solely on contrastive learning is insufficient to replicate the performance of a binary classifier that is free from inversely ordered pairs. Therefore, it is reasonable to consider the optimal binary classifier as the ultimate state of contrastive learning models.

\subsection{Spearman's Correlation Coefficient}

Before deriving the performance upper bound of contrastive learning methods on STS tasks, it is essential to introduce Spearman's correlation coefficient, the primary evaluation metric in this field. This statistic measures the ordinal consistency between the cosine similarity of embeddings and human ratings, as defined by Equation~\ref{eq:spearman}:
\begin{equation}
    \rho = 1 - \frac{6 \sum d_i^2}{n(n^2 - 1)}
\label{eq:spearman}
\end{equation}

In this formula, $n$ represents the number of data points, and $d_i$ is the difference between the rank of the $i$-th sentence pair's cosine similarity after encoding into embeddings and its human-judged similarity rank. Particularly, when multiple entries share the same rating, their ranks are substituted with their mean rank during the computation of Equation~\ref{eq:spearman}.

Spearman's correlation coefficient, ranging from $[-1, 1]$, indicates stronger consistency between model outputs and human evaluations as it approaches 1. Typically, the coefficient is multiplied by 100 to yield a percentage score, facilitating more straightforward comparisons of encoding effectiveness across different models. 

\subsection{The Spearman Correlation Upper Limit of Contrastive Learning Methods}
\label{sec:upper_bound}

As discussed in section~\ref{sec:binary}, contrastive learning distinguishes texts based on coarse-grained semantic relations, categorizing them as either similar or dissimilar. Thus, its effectiveness parallels that of a binary classifier. This section derives the optimal Spearman correlation achievable by a binary classifier in STS tasks, thereby elucidating the performance upper bound of contrastive learning methods.

Given a collection of text pairs $\mathrm{X}=\{(x_i, x_i^?)\}_1^n$ consisting of $n$ samples, we initially arrange the elements of $\mathrm{X}$ in descending order according to manually annotated semantic similarity, yielding the sorted set $\mathrm{Y}=\{(y_i, y_i^?)\}_1^n$. Assume that $\cos(y_{k}, y_{k}^?) > \cos(y_{k+1}, y_{k+1}^?), \forall k\in [1, n-1]$. Then, for any binary classifier, its performance reaches the optimum only when it categorizes the first $k$ sample pairs of $\mathrm{Y}$ as positive examples and the remaining $n-k$ sample pairs as negatives. Otherwise, it indicates at least one misclassification.

Since this binary classifier is solely responsible for constructing an optimal classification boundary between the two categories of similarity and dissimilarity (i.e., determining only whether two texts are semantically akin), its predicted scores for the first $k$ samples are consistently identical (assumed to be 1), and likewise for the last $n-k$ samples (assumed to be 0). By the definition of Spearman's correlation coefficient, the difference in rankings between predictions and true values, $d_i$, alongside $\sum d_i^2$, can be represented as:
\begin{equation}
\resizebox{\linewidth}{!}{$
\begin{aligned}
d_i &= i - \frac{k+1}{2}, \quad i = 1, 2, \ldots, k\\
d_i &= i - \frac{k+n+1}{2}, \quad i = k+1, k+2, \ldots, n\\
\sum d_i^2 &= \sum_{i=1}^{k} (i - \frac{k+1}{2})^2 + \sum_{i=k+1}^{n} (i - \frac{k+n+1}{2})^2
\end{aligned}
$}
\label{eq:d_i}
\end{equation}

These equations showcase that $\sum d_i^2$ can be viewed as a function of $k$. Upon rearranging, we derive: (with further details provided in Appendix~\ref{appendix:detail}.)
\begin{equation}
\resizebox{\linewidth}{!}{$
\begin{aligned}
\sum d_i^2 &= \sum_{i=1}^{k} (i - \frac{k+1}{2})^2 + \sum_{i=k+1}^{n} (i - \frac{k+n+1}{2})^2 \\
&= \sum_{i=1}^{k} (i - \frac{k+1}{2})^2 + \sum_{i=k+1}^{n}\Big((i-\frac{k+1}{2}) -\frac{n}{2} \Big)^2 \\
&= \sum_{i=1}^n (i - \frac{k+1}{2})^2 + (n-k)\frac{n^2}{4} -n\sum_{i=k+1}^{n}(i-\frac{k+1}{2}) \\
&= \sum_{i=1}^{n}i^2 +\frac{n(k+1)^2}{4} -\frac{n(n+1)(k+1)}{2} - \frac{n^2(n-k)}{4} \\
&= \frac{n(n+1)(2n+1)}{6} + \frac{n}{4}\Big(k^2 -nk -(n+1)^2\Big)
\end{aligned}
$}
\label{eq:sum_d_i}
\end{equation}

In Equation~\ref{eq:sum_d_i}, $n$ remains constant, so $\sum d_i^2$ depends on $f(k)=k^2 - nk -(n+1)^2$. When $k = \frac{n}{2}$, i.e., when the model deems the first $50\%$ of sample pairs as positives and the remaining $50\%$ as negatives, $f(k)$ attains its minimum. Therefore, the minimum value of $\sum d_i^2$ is:
\begin{equation}
\resizebox{\linewidth}{!}{$
\begin{aligned}
\min\Big(k^2 -nk -(n+1)^2 \Big) = &-\frac{5n^2}{4} -2n - 1 \\
\min (\sum d_i^2) = \frac{n(n+1)(2n+1)}{6} &-\frac{n}{4}(\frac{5n^2}{4} +2n + 1)
\end{aligned}
$}
\label{eq:min_d_i}
\end{equation}

Subsequently, by substituting $\min(\sum d_i^2)$ into the expression for Spearman's correlation coefficient (Equation~\ref{eq:spearman}), the maximum Spearman correlation achievable by this binary classifier is 0.875. This indicates that the optimal performance of contrastive learning in STS tasks will not exceed 0.875.
\begin{equation}
\begin{aligned}
\max (\rho) &= 1 - \frac{{n^2} - 4}{8(n^2 - 1)} = \frac{7n^2-4}{8(n^2-1)}\\
\lim_{n\to\infty}\max (\rho) &= \lim_{n\to\infty}\frac{7n^2-4}{8(n^2-1)} = \frac{7}{8} = 0.875
\end{aligned}
\label{eq:max_spearman}
\end{equation}

Apart from the original InfoNCE Loss, an extended contrastive learning loss function tailored for NLI datasets \citep{snli-2015-EMNLP, mnli-2018-NAACL}, as shown in Formula~\ref{eq:extend_info_nce}, is frequently utilized in sentence representation research \citep{SimCSE-EMNLP-2021, CoT-BERT-ICANN-2024}. The incorporation of hard negative example $x_j^-$ in the denominator, which is equivalent to enlarging the batch size, does not affect the correctness of our derivation.
\begin{equation}
\resizebox{\linewidth}{!}{$
- {\log} \frac{e^{{\cos}(f(x_i),f(x_i^+))/\tau}}{\sum_{j=1}^N\big(e^{{\cos}(f(x_i),f(x_j^+))/\tau} + e^{{\cos}(f(x_i),f(x_j^-))/\tau}\big)}
$}
\label{eq:extend_info_nce}
\end{equation}

It should be noted that the above conclusion has been validated through numerous experiments. To date, embedding derivation schemes based on contrastive learning have not achieved a Spearman's correlation score above 86.5. This theoretical analysis provides a clear explanation for these empirical observations.

\section{Proposed Method}
\label{sec:method}

This section introduces Pcc-tuning, an innovative strategy for addressing STS tasks. Pcc-tuning employs a two-stage training pipeline and is designed to break the 87.5 performance ceiling in contrastive learning methods.

The anisotropy of PLMs' semantic space \citep{Anisotropy-EMNLP-2019} is a longstanding challenge in sentence representation research. Contrastive learning has proven effective in stabilizing embedding distances among semantically similar texts while promoting a more uniform distribution of overall vector encodings \citep{SimCSE-EMNLP-2021}, thus markedly enhancing the semantic properties of PLMs. Therefore, leveraging contrastive learning to refine the initial state of pre-trained models has emerged as a prevalent approach within the NLP community \citep{E5-2022, GTE-2023, GRIT-2024}.
\begin{figure*}[ht]
\centering
\includegraphics[width=1.0\linewidth]{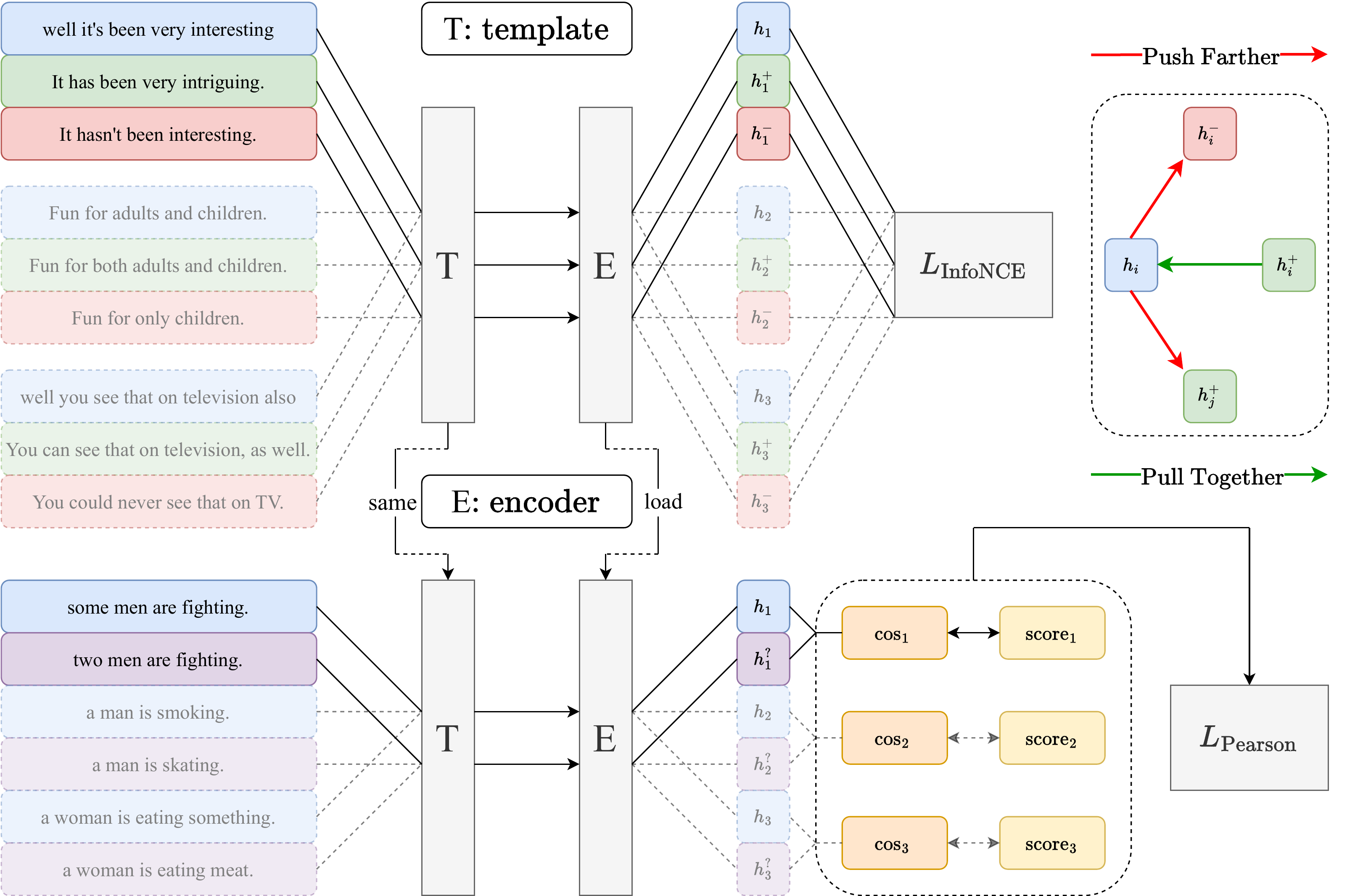}
\caption{The overall architecture of Pcc-tuning. By default, we use "\textit{This sentence : `[X]' can be summarized as}" \citep{PretCoTandKE-ICIC-2024} as the manual template for both stages. In the diagram, $h_i$ denotes the embedding of sentence $s_i$ after model encoding, $\cos_i$ represents the cosine similarity between $h_i$ and $h_i^?$, while $\text{score}_i$ is the human-annotated similarity score for $s_i$ and $s_i^?$.}
\label{fig:workflow}
\end{figure*}

Following this well-established practice, we initially conduct supervised fine-tuning of the PLM using the NLI dataset constructed by SimCSE \citep{SimCSE-EMNLP-2021}. This dataset comprises 275,601 triplet-form text pairs, providing a robust source of coarse-grained labeled information for the model. Our implementation in the first stage closely mirrors that of PromptEOL \citep{PromptEOL-2023} for comparison purposes, where we load the original PLM checkpoint and fine-tune the model with the extended InfoNCE Loss depicted in Equation~\ref{eq:extend_info_nce}, combined with QLoRA \citep{QLoRA-NIPS-2024}. A distinctive feature of our methodology is the adoption of the PromptSUM template proposed by \citet{PretCoTandKE-ICIC-2024}: "\textit{This sentence : `[X]' can be summarized as}", which encapsulates the input sentence [X] and extracts the encoding of the final token as the sentence embedding. Later sections will examine Pcc-tuning's performance under various prompts.

After the contrastive learning phase, the PLM will be adjusted to a superior encoding state, capable of producing high-quality embeddings. However, the neural network trained at this stage remains insufficient as the final solution for STS tasks. This is due to two primary reasons: (1) The contrastive learning objective does not fully align with the evaluation metrics of STS tasks. While a decrease in InfoNCE Loss reflects a better clustering effect of sentence vectors in semantic space, this does not necessarily translate into an improvement in Spearman correlation. The latter essentially measures the consistency of the model's scoring with human ratings in terms of monotonicity. (2) As discussed in section~\ref{sec:proof}, contrastive learning loss functions fail to harness fine-grained annotation information, leading to a pronounced performance bottleneck. Consequently, the benefits of further optimizing binary classification performance diminish with successive iterations. To mitigate these issues, a finer distinction is required within the two categories of similarity and dissimilarity, along with introducing ordinal relationships of text pairs based on semantic similarity.

The optimal strategy is to incorporate fine-grained annotated data in the second stage and guide the model's training process via Spearman's correlation coefficient. This ensures maximum consistency between the model's behavior during training and testing. However, since Spearman correlation is non-differentiable and thus incompatible with backpropagation, we opt for Pearson's correlation coefficient to update model parameters, which also serves as the inspiration for the name Pcc-tuning. Pearson correlation and our loss function for the second stage are shown in Equation~\ref{eq:pearson}, where $\mathrm{X}$ represents the cosine similarity between model-derived embeddings, and $\mathrm{Y}$ denotes the human-annotated scores for the text pairs.
\begin{equation}
\begin{aligned}
r &= \frac{\text{cov}(\mathrm{X}, \mathrm{Y})}{\sigma_\mathrm{X} \sigma_\mathrm{Y}} \\
\ell_{p} &= - r + 1 \in [0, 2]
\end{aligned}
\label{eq:pearson}
\end{equation}

Concretely, for a batch of text pairs $\{(x_i, x_i^?)\}_1^N$, we first invoke the PLM to encode $x_i$ and $x_i^?$, obtaining $f(x_i)$ and $f(x_i^?)$. Then, we directly compute their cosine similarity and store the result in $\mathrm{X}=\{\cos(f(x_i), f(x_i^?))\}_1^N$. Subsequently, we input $\mathrm{X}$ and the true similarity scores $\mathrm{Y}=\{y_i\}_1^N$ into Equation~\ref{eq:pearson} to calculate the loss.

Employing Pearson's correlation coefficient as the loss function enables effective utilization of fine-grained sample scores and supports diverse combinations even with relatively small data quantities. For example, our tuning dataset in the second stage is composed of filtered training sets from STS-B \citep{STS-B-2017} and SICK-R \citep{SICK-R-2014}, which together contain 5,398 text pairs. This number merely constitutes 1.96\% of the size of the NLI dataset adopted in the first stage, yet the potential combination varieties reach up to $\text{C}_{5398}^N$ (where $N$ represents the batch size). As a result, even after multiple epochs of training, the similarity rankings of samples in each batch are unlikely to repeat, thereby continuously providing the model with meaningful gradient information.

The dataset is filtered because we discovered that some sentence pairs in the STS-B and SICK-R training sets overlap with the test sets of the seven STS benchmarks in SentEval. Additionally, there is even overlap between the train and test sets within SICK-R itself. To prevent information leakage, we implemented a stringent filtering mechanism to ensure that the model does not encounter any test set text pairs during parameter updates. More details about this filtering process can be found in Appendix~\ref{appendix:filter}.

Figure~\ref{fig:workflow} presents an overview of Pcc-tuning's training pipeline. In the first stage, the model is fine-tuned using contrastive learning on the NLI corpus. In the second stage, we introduce a small amount of fine-grained annotated data and load the checkpoint from the first phase to further update the model parameters via Pearson's correlation coefficient. This two-stage fine-tuning strategy effectively prevents overfitting. Although the filtered STS-B and SICK-R training sets provide only 5,398 fine-grained labeled instances, the 275,601 text pairs used in the first stage establish a solid initial state for the model, thereby maximizing its generalization capacity. Moreover, the varying scoring scales of STS-B and SICK-R also introduce a degree of noise into the Pcc-tuning training process , which contributes to the model's robustness. We provide detailed results of our method's zero-shot performance on several downstream tasks in Appendix~\ref{appendix:transfer} to further demonstrate its transferability.

\section{Experiments}
\label{sec:exp}

This section presents the experimental results of Pcc-tuning. Initially, in subsection~\ref{sec:details}, we elaborate on our experimental setup, including evaluation methods, training data size, and the selection of baselines. Subsequently, in subsection~\ref{sec:main_results}, we compare the performance of Pcc-tuning with current SOTA text representation strategies across internationally recognized STS benchmarks. Following this, in subsection~\ref{sec:contrast}, we conduct targeted experiments to demonstrate that Pcc-tuning surpasses continuous contrastive learning. Finally, in subsection~\ref{sec:prompts}, we examine the efficacy of Pcc-tuning under different prompts.

\subsection{Implementation Details}
\label{sec:details}

In line with prior studies \citep{SimCSE-EMNLP-2021, PromptBERT-EMNLP-2022, PromptEOL-2023, SSCL-ACL-2023}, we utilize the SentEval \citep{SentEval-LREC-2018} toolkit to assess our model on seven STS tasks, with Spearman's correlation coefficient as the core metric. In all experiments, models are permitted access to text pairs from the evaluation benchmarks only during the testing phase. 

It is noteworthy that although Pcc-tuning requires specific corpora at both stages of training, the total data volume employed is only 280,999 entries. In contrast, the publicly available training data for the contemporary SOTA method, DeeLM \citep{DeeLM-2023}, includes 480,862 triplet text pairs, with additional datasets remaining inaccessible.
\begin{table*}[ht]
\centering
\resizebox{1.0\linewidth}{!}{
    \begin{tabular}{ccccccccc}
    \toprule 
    \bf{Methods} & \bf{STS-12} & \bf{STS-13} & \bf{STS-14} & \bf{STS-15} & \bf{STS-16} & \bf{STS-B} & \bf{SICK-R} & \bf{Avg.} \\
    \midrule \midrule
    \multicolumn{9}{c}{\textbf{Pre-trained Embedding Models}} \\
    \midrule
    openai-ada-002 $^\S$ & 69.80 & 83.27 & 76.09 & 86.12 & 85.96 & 83.17 & 80.60 & 80.72 \\
    jina-base-v2 $^\ddagger$ & 74.28 & 84.18 & 78.81 & 87.55 & 85.35 & 84.85 & 78.98 & 82.00 \\
    nomic-embed-v1 $^\ddagger$ & 73.75 & 85.03 & 80.52 & 87.40 & 83.55 & 83.90 & 76.52 & 81.52 \\
    \midrule \midrule
    \multicolumn{9}{c}{\textbf{Fine-tuning Strategies}} \\
    \midrule
    \multicolumn{9}{c}{{\textbf{Previous SOTA methods}}. \it{Implementation on $\rm{LLaMA2}_{\rm 7b}$}} \\
    SimCSE $^\diamondsuit$ & 78.39 & 89.95 & 84.80 & 88.50 & 86.04 & 87.86 & 81.11 & 85.24 \\
    PromptEOL & 79.24 & 90.31 & 84.74 & 88.72 & 86.01 & 87.87 & 80.94 & 85.40 \\
    AnglE $^\diamondsuit$ & 79.00 & 90.56 & 85.79 & 89.43 & 87.00 & 88.97 & 80.94 & 85.96 \\
    DeeLM $^\diamondsuit$ & 79.01 & 90.32 & 85.84 & 89.47 & 87.18 & 89.15 & 81.08 & 86.01 \\
    \midrule
    \multicolumn{9}{c}{\it{Implementation on $\rm{OPT}_{\rm 6.7b}$}} \\
    PromptSUM & 79.66 & 89.91 & 84.96 & 89.60 & 85.79 & 88.54 & 80.51 & 85.57 \\
    Pcc-tuning & \bf 80.04 & \bf 90.41 & \bf 85.63 & \bf 90.53 & \bf 86.32 & \bf 89.37 & \bf 86.21 & \bf 86.93 \\
    \midrule
    \multicolumn{9}{c}{\it{Implementation on $\rm{LLaMA}_{\rm 7b}$}} \\
    PromptSUM & 78.84 & 90.03 & 85.06 & 88.80 & 85.66 & 88.29 & 81.58 & 85.47 \\
    Pcc-tuning (SUM) & \bf 81.00 & \bf 90.66 & \bf 86.09 & \bf 90.42 & \bf 86.21 & \bf 89.83 & \bf 87.23 & \bf 87.35  \\
    PromptEOL & 79.00 & 89.80 & 85.10 & 88.86 & 86.03 & 88.48 & 81.06 & 85.48 \\
    Pcc-tuning (EOL) & \bf 81.41 & \bf 91.15 & \bf 86.62 & \bf 90.69 & \bf 86.99 & \bf 89.97 & \bf 86.85 & \bf 87.67 \\
    \midrule
    \multicolumn{9}{c}{\it{Implementation on $\rm{LLaMA2}_{\rm 7b}$}} \\
    PromptSUM & 79.43 & 90.25 & 85.03 & 88.71 & 86.07 & 87.96 & 81.28 & 85.53 \\
    Pcc-tuning & \bf 81.82 & \bf 91.36 & \bf 86.88 & \bf 90.66 & \bf 87.04 & \bf 89.73 & \bf 87.11 & \bf 87.80 \\
    \midrule
    \multicolumn{9}{c}{\it{Implementation on $\rm{Mistral}_{\rm 7b}$}} \\
    PromptSUM & 79.76 & 89.69 & 85.33 & 89.30 & 86.62 & 88.27 & 81.81 & 85.83 \\
    Pcc-tuning & \bf 82.04 & \bf 90.84 & \bf 86.79 & \bf 91.10 & \bf 87.18 & \bf 90.05 & \bf 87.02 & \bf 87.86 \\
    \bottomrule
    \end{tabular}
}
\caption{Spearman’s correlation scores across seven STS benchmarks for different methods. This table highlights Pcc-tuning's comprehensive two-stage training strategy in comparison with PromptSUM / EOL, which corresponds to the first stage of Pcc-tuning. Please refer to section~\ref{sec:prompts} for the specific structure of PromptEOL and PromptSUM. $\S$: results from \citep{MTEB-2022}. $\ddagger$: results from \citet{STS-Regression-2024}. $\diamondsuit$: results from \citep{DeeLM-2023}.} 
\label{tab:main_results}
\end{table*}

Our experiments are conducted based on several widely adopted 7B-scale generative PLMs: OPT$_{\rm 6.7b}$ \citep{OPT-2022}, LLaMA$_{\rm 7b}$ \citep{LLaMA-2023}, LLaMA2$_{\rm 7b}$, and Mistral$_{\rm 7b}$. To clearly demonstrate the superiority of Pcc-tuning, we primarily compare it against current SOTA strategies. Specifically, among our selected baselines, PromptEOL \citep{PromptEOL-2023}, PromptSUM \citep{PretCoTandKE-ICIC-2024}, AngIE \citep{AnglE-2023}, and DeeLM \citep{DeeLM-2023} are leading generative PLM-based sentence representation methods, which significantly outperform BERT-based approaches on STS benchmarks. Meanwhile, openai-ada-002, jina-base-v2 \citep{Jina2-2023}, and nomic-embed-v1 \citep{NomicEmbed-2024} represent the most advanced contrastive learning pre-trained models at present.

\subsection{Main Results}
\label{sec:main_results}

Table~\ref{tab:main_results} summarizes the performance of various methods on the seven STS tasks collected in SentEval. Under all tested PLMs, Pcc-tuning consistently surpasses previous SOTA strategies, either approaching or exceeding the Spearman correlation upper limit of 87.5 for contrastive learning methods. Notably, when Mistral$_{\rm 7b}$ is selected as the backbone, Pcc-tuning attains a Spearman correlation of 87.86, which is 2.15\% higher than the leaderboard record of 86.01 set by DeeLM, despite DeeLM using a much larger training corpus. Moreover, considering that Pcc-tuning delivers the best performance across all seven STS tasks, its effectiveness is self-evident. These outcomes collectively underscore the crucial role of modeling fine-grained annotated information in STS tasks.

Furthermore, since Pcc-tuning's first-stage is implemented identically to PromptSUM, the comparison between Pcc-tuning and PromptSUM in Table~\ref{tab:main_results} also functions as an ablation study. It reveals that, constrained by the coarse granularity of contrastive learning, whether adopting the earlier released OPT model or the newly open-sourced Mistral model, Spearman's correlation scores for PromptSUM are confined around 85.5, showing limited progress. In contrast, Pcc-tuning provides an improvement of approximately 2 percentage points, reaffirming the mathematical derivations discussed in section~\ref{sec:proof}.

In addition to the inability in fully harnessing fine-grained annotated data, another significant drawback of contrastive learning is its reliance on large batch sizes to  ensure negative sample diversity, which consumes substantial computational resources \citep{PromptEOL-2023, PretCoTandKE-ICIC-2024}. To explore Pcc-tuning's memory consumption and the impact of batch size on model performance, we conducted relevant experiments detailed in Appendix~\ref{appendix:batch}. The findings indicate that Pcc-tuning demonstrates superior memory efficiency and exhibits strong robustness to varying batch sizes.

\subsection{Pcc-tuning vs. Two-Stage Contrastive Learning}
\label{sec:contrast}

This section addresses an intriguing question: Can contrastive learning, when supplemented with fine-grained annotated data, further enhance model performance? Specifically, when employing the filtered STS-B and SICK-R training sets for two-stage parameter updates, does Pcc-tuning outperform continuous contrastive learning?

As analyzed in section~\ref{sec:proof}, contrastive learning methods are limited in their ability to fully leverage fine-grained annotated data. Therefore, to apply the STS-B and SICK-R training sets to contrastive learning models, a threshold must be selected to identify suitable positive sample pairs, which inevitably leads to significant data loss. After balancing dataset scale and quality, we chose to treat text pairs with similarity scores greater than 4.0 as positive samples. Following this step, the original 5,398 data pairs were reduced to 1,543.

Table~\ref{tab:vs} presents the results of the above experiments. In the second column, "Contrastive" and "Pearson" refer to fine-tuning the first-stage checkpoint using either contrastive learning or Pcc-tuning, respectively. The "Performance" column reports the model’s average Spearman's correlation scores across seven STS benchmarks. The results clearly show that continuing with contrastive learning not only yields significantly inferior results compared to Pcc-tuning, but also underperforms the model’s first-stage outcomes.
\begin{table}[htbp]
\centering
\begin{tabular}{c|c|c}
    \toprule
    \bf PLMs & \bf Strategy  & \bf Performance \\
    \midrule
    \multirow{3}{*}{OPT$_{\rm 6.7b}$} 
    & Stage I & 85.57 \\ 
    & Contrastive & 77.29 \\
    & Pearson & \bf 86.93 \\
    \midrule
    \multirow{3}{*}{LLaMA$_{\rm 7b}$} 
    & Stage I & 85.47 \\
    & Contrastive & 79.47 \\
    & Pearson & \bf 87.35 \\
    \midrule
    \multirow{3}{*}{LLaMA2$_{\rm 7b}$}
    & Stage I & 85.53 \\
    & Contrastive & 85.38 \\
    & Pearson & \bf 87.80 \\
    \midrule
    \multirow{3}{*}{Mistral$_{\rm 7b}$} 
    & Stage I & 85.83 \\
    & Contrastive & 75.47 \\
    & Pearson & \bf 87.86 \\
    \bottomrule
\end{tabular}
\caption{Performance comparison between Pcc-tuning and two-stage contrastive learning strategies on STS benchmarks.}
\label{tab:vs}
\end{table}

These findings are not surprising, as contrastive learning methods require large batch sizes to avoid model collapse. Indeed, most mainstream contrastive learning-based text representation models employ batch sizes of 256 or more. In comparison, such a small amount of annotated data is hardly sufficient to effectively support contrastive learning. However, it is important to note that due to the coarse-grained semantic partitioning inherent in InfoNCE Loss, even with a larger corpus, contrastive learning methods cannot surpass Pcc-tuning. This is because Pcc-tuning possess higher data utilization efficiency and is more closely aligned with the evaluation metrics of STS tasks.

\subsection{Pcc-tuning under Various Prompts}
\label{sec:prompts}

The Explicit One-word Limitation (EOL) template, introduced by PromptEOL \citep{PromptEOL-2023}, represents a pioneering effort in employing generative PLMs for embedding derivation and has become the most widely adopted prompt in sentence representation research. Recently, \citet{PretCoTandKE-ICIC-2024} proposed two alternative templates, PromptSTH and PromptSUM, which deviate from the EOL structure. Their findings demonstrated that strict adherence to the EOL format is not necessary for effective PLM fine-tuning. The specific forms of these prompts are depicted in Table~\ref{tab:prompts}, where [X] represents the input text, and the parts highlighted in red denote the positions from which the model extracts embeddings.
\begin{table}[htbp]
\renewcommand \arraystretch{1.3}
\centering
\begin{tabular}{c}
\toprule
{\bf PromptEOL}  \\
This sentence : "[X]" means in one word:\textcolor{red}{"} \\
\midrule 
{\bf PromptSUM} \\
This sentence : "[X]" can be summarized \textcolor{red}{as} \\
\midrule 
{\bf PromptSTH} \\
This sentence : "[X]" means \textcolor{red}{something} \\
\bottomrule 
\end{tabular}
\caption{Manual templates employed by PromptEOL, PromptSUM, and PromptSTH. Apart from differences in prompts, the implementations of these three methods are completely identical.}
\label{tab:prompts}
\end{table}

To further validate the versatility of our approach, we assessed the average Spearman's correlation scores on seven STS tasks using these prompts as the templates for both stages of Pcc-tuning. The corresponding results are delineated in Table~\ref{tab:prompts-sts}. As evidenced by the results, Pcc-tuning consistently improves model performance, with minimal impact from the different templates on the final outcomes. This suggests that when applying Pcc-tuning to downstream tasks, there is little need for laborious prompt searches, thereby offering significant practical benefits.
\begin{table}[htbp]
\centering
\resizebox{1.0\linewidth}{!}{
\begin{tabular}{c|c|c|c}
    \toprule
    \bf PLMs & \bf Templates & \bf Stage I & \bf Stage II \\
    \midrule
    \multirow{3}{*}{OPT$_{\rm 6.7b}$} 
    & PromptSTH & 85.51 & 86.86 \\
    & PromptEOL & 85.52 & \bf 86.96 \\
    & PromptSUM & 85.57 & 86.93 \\
    \midrule
    \multirow{3}{*}{LLaMA$_{\rm 7b}$} 
    & PromptSTH & 85.40 & 87.54 \\
    & PromptEOL & 85.48 & \bf 87.67 \\
    & PromptSUM & 85.47 & 87.35 \\
    \midrule
    \multirow{3}{*}{LLaMA2$_{\rm 7b}$} 
    & PromptSTH & 85.31 & 87.64 \\
    & PromptEOL & 85.40 & 87.75 \\
    & PromptSUM & 85.53 & \bf 87.80 \\
    \midrule
    \multirow{3}{*}{Mistral$_{\rm 7b}$} 
    & PromptSTH & 85.66 & 87.70 \\
    & PromptEOL & 85.50 & 87.72 \\
    & PromptSUM & 85.83 & \bf 87.86 \\
    \bottomrule
\end{tabular}
}
\caption{Average Spearman's correlation scores obtained by Pcc-tuning on seven STS benchmarks under different PLMs and manual templates. The settings for stage I and stage II are consistent with the descriptions in section~\ref{sec:method}.}
\label{tab:prompts-sts}
\end{table}

\section{Related Work}

Contrastive learning is currently the principal strategy within the NLP community for addressing STS tasks, and our proposed method, Pcc-tuning, is specifically designed to overcome the inherent limitations of these approaches.

Prior to the rise of contrastive learning-based text representation schemes, Sentence-BERT had already introduced the idea of enhancing the semantic encoding capabilities of PLMs using the STS-B training set \citep{Sentence-BERT-EMNLP-2019}. However, subsequent contrastive learning methods, such as SimCSE \citep{SimCSE-EMNLP-2021}, PromptBERT \citep{PromptBERT-EMNLP-2022}, and CoT-BERT \citep{CoT-BERT-ICANN-2024} have demonstrated superior performance across the seven STS benchmarks collected in SentEval, thereby making them the focal point of recent academic research and development. 

To the best of our knowledge, this study is the first to propose and substantiate the theoretical performance upper bound of contrastive learning methods. Additionally, Pcc-tuning is the inaugural method capable of achieving Spearman's correlation scores above 87 on standard STS tasks, marking a significant advancement in the field.

\section{Conclusion}

In this paper, we first analyze the structure of contrastive learning loss functions, highlighting that their coarse-grained categorization of semantic relationships among texts renders contrastive learning akin to a binary classifier. Building on this insight, we rigorously derive the optimal Spearman correlation achievable by a binary classifier in STS tasks, establishing that the upper bound for the Spearman correlation of contrastive learning methods is 87.5. To achieve further breakthroughs, we introduce Pcc-tuning, a novel strategy that effectively harnesses fine-grained annotated information. Pcc-tuning leverages a two-stage training pipeline and utilizes Pearson's correlation coefficient as the loss function to fully exploit the ordinal relationships between text pairs. Extensive experimental results demonstrate that Pcc-tuning significantly enhances the quality of generated embeddings, with consistent performance gains observed across various PLMs, prompts, and batch sizes.

\section*{Limitations}

In preparing the training dataset for the second stage of Pcc-tuning, we employ a mixed corpus composed of the training sets from STS-B and SICK-R, while filtering out any overlapping samples with the test sets. However, the labeling scales of these two datasets are not congruent. Specifically, the similarity scores in the STS-B training set span from 0 to 5, whereas the scores in the SICK-R training set range from 1 to 5. To unify their annotation scales, we transform each label in the SICK-R training set using the formula $5 \times \frac{\text{label} - 1}{4}$, thereby converting the scores to the range of $[0, 5]$. Given that this is merely a simple linear mapping, it is likely that some vital manually annotated information is lost, potentially hindering Pcc-tuning from reaching its optimal performance on the evaluation benchmarks.

\bibliography{custom}

\appendix

\section{Derivation Details}
\label{appendix:detail}

Due to space constraints, some steps in the calculation are abbreviated when rearranging Equation~\ref{eq:sum_d_i} in section~\ref{sec:upper_bound}. Here, we provide the complete derivation process:
\begin{equation}
\resizebox{\linewidth}{!}{$
\begin{aligned}
& \sum d_i^2 \\
    &= \sum_{i=1}^{k} (i - \frac{k+1}{2})^2 + \sum_{i=k+1}^{n} (i - \frac{k+n+1}{2})^2 \\
     &= \sum_{i=1}^{k} (i - \frac{k+1}{2})^2 + \sum_{i=k+1}^{n}\Big((i-\frac{k+1}{2}) -\frac{n}{2} \Big)^2 \\
     &= \sum_{i=1}^{k} (i - \frac{k+1}{2})^2 + \sum_{i=k+1}^{n}\Big((i-\frac{k+1}{2})^2+\frac{n^2}{4} - n(i-\frac{k+1}{2}) \Big) \\
     &= \sum_{i=1}^n (i - \frac{k+1}{2})^2 + \sum_{i=k+1}^{n}\Big(\frac{n^2}{4} - n(i-\frac{k+1}{2})\Big) \\
     &= \sum_{i=1}^n (i - \frac{k+1}{2})^2 + (n-k)\frac{n^2}{4} -n\sum_{i=k+1}^{n}(i-\frac{k+1}{2}) \\
     &= \sum_{i=1}^n (i - \frac{k+1}{2})^2 + (n-k)\frac{n^2}{4} -n\Big(\frac{(n-k)(n+k+1)}{2}-\frac{(n-k)(k+1)}{2} \Big) \\
     &= \sum_{i=1}^n (i - \frac{k+1}{2})^2 + (n-k)\frac{n^2}{4} -n(n-k)\Big(\frac{(n+k+1)}{2}-\frac{(k+1)}{2} \Big) \\
     &= \sum_{i=1}^n (i - \frac{k+1}{2})^2 + \frac{n^2(n-k)}{4} - \frac{n^2(n-k)}{2} \\
     &= \sum_{i=1}^n (i - \frac{k+1}{2})^2 - \frac{n^2(n-k)}{4} \\
     &= \sum_{i=1}^{n}\Big(i^2 + \frac{(k+1)^2}{4} -(k+1)i \Big) - \frac{n^2(n-k)}{4}\\
     &= \sum_{i=1}^{n}i^2 +\frac{n(k+1)^2}{4} -\frac{n(n+1)(k+1)}{2} - \frac{n^2(n-k)}{4}\\ 
    &=\frac{n(n+1)(2n+1)}{6} + \frac{n}{4}\Big((k+1)^2 -2(k+1)(n+1)-n(n-k)\Big)\\
    &= \frac{n(n+1)(2n+1)}{6} + \frac{n}{4}\Big(k^2+2k+1 -2(n+1)-2(n+1)k -n^2+nk\Big)\\
    &= \frac{n(n+1)(2n+1)}{6} + \frac{n}{4}\Big(k^2 -nk -(n+1)^2\Big)
\end{aligned}
$}
\end{equation}

\section{Data Filtering Method}
\label{appendix:filter}

In section~\ref{sec:method}, we mentioned that the fine-tuning data in the second stage of Pcc-tuning is derived from filtered STS-B and SICK-R training sets, aimed at preventing the model from encountering text pairs present in the evaluation benchmarks during parameter updates. Here, we provide a detailed description of the filtering process.

In standard STS tasks, each sample consists of two text strings, \( s \) and \( s^? \), along with a floating-point number \( \text{gs} \) indicating the semantic similarity score between them. In our experimental setup, for any sample \((s_1, s_1^?, \text{gs}_1)\) from the STS-B or SICK-R training sets, if a text pair \((s_2, s_2^?, \text{gs}_2)\) exists in the test sets of STS12-16, STS-B, or SICK-R, where \(s_1 = s_2\) and \(s_1^? = s_2^?\), or \(s_1 = s_2^?\) and \(s_1^? = s_2\), we treat them as duplicates, regardless of whether \(\text{gs}_1\) and \(\text{gs}_2\) are identical. All duplicate training samples are then removed from the model’s fine-tuning corpus.

This process involves a highly stringent filtering mechanism. Under this approach, even the train and test sets of SICK-R itself contain overlapping samples (despite differences in their \(\text{gs}\) values). The original STS-B and SICK-R training sets consist of 5,749 and 4,500 samples, respectively. After filtering, they are reduced to 991 and 4,407 samples, respectively, resulting in a total of 5,398 text pairs. Moreover, as noted in the Limitations section of this paper, the annotation scales of these two datasets are not consistent, which leads to some information loss during the merging process. The scarcity of annotated data, combined with differences in scoring standards, posed additional challenges for this study.

Despite these challenges, Pcc-tuning still demonstrated strong performance (section~\ref{sec:exp}, Appendix~\ref{appendix:batch}). This suggests that in downstream scenarios with more abundant task-specific data, the advantages of Pcc-tuning over contrastive learning could become even more pronounced, highlighting its broad potential for application.

\section{Transfer Tasks}
\label{appendix:transfer}

In the previous sections, we have thoroughly validated the exceptional performance of Pcc-tuning in semantic matching across seven well-established STS benchmarks. To further clarify its generalization capabilities, this section evaluates the transferability of Pcc-tuning through zero-shot testing on a variety of downstream tasks.
\begin{table*}[ht]
    \centering
    \begin{tabular}{|c|c|c|c|} 
    \hline
    \diagbox{Task}{Method} & LLaMA2 (raw) & PromptSUM & Pcc-tuning \\
    \hline
    Banking77Classification & 56.38 & 85.41 & \textbf{86.09} \\
    TwitterSemEval2015      & 79.05 & 87.66 & \textbf{87.73} \\
    AskUbuntuDupQuestions  & 41.46 & 58.00 & \textbf{61.19} \\
    StackOverflowDupQuestions   & 24.44 & 44.93 & \textbf{47.65} \\
    CQADupstackEnglishRetrieval   & 0.17 & 32.31 & \textbf{34.19} \\
    LegalSummarization   & 7.49 & 66.20 & \textbf{68.31} \\
    FaithDial & 0.52 & 25.54 & \textbf{28.28} \\ 
    PIQA & 1.06 & 30.81 & \textbf{31.07} \\
    \hline
    \end{tabular}
    \caption{Model performance on eight downstream tasks. The reported values represent the primary evaluation metric for each task, scaled by 100 to convert them into percentage scores.}
    \label{tab:down_stream}
\end{table*}

We selected eight tasks of different types from the Massive Text Embedding Benchmark (MTEB) \citep{MTEB-2022} to cover four major areas: Retrieval, Reranking, Classification, and Pair Classification. For these experiments, we employed LLaMA2$_{\rm 7b}$ as the backbone and directly loaded model checkpoints without any additional parameter updates. In other words, the PromptSUM and Pcc-tuning models used here are identical to those in Table~\ref{tab:main_results}.

Table~\ref{tab:down_stream} summarizes the results of these tests. In this table, "LLaMA2 (raw)" refers to the original LLaMA2 model without any integrated prompts. It can be observed that the transferability of raw LLaMA2 output vectors is quite poor, with scores on some tasks falling below 1\%. This suggests a substantial gap between auto-regressive language modeling and effective embedding generation. However, the scores for PromptSUM demonstrate a significant improvement over LLaMA2 (raw), indicating that contrastive learning can enhance the embedding quality of generative PLMs. Pcc-tuning further amplifies this effect. Despite introducing only an additional 5,398 fine-grained annotated text pairs, Pcc-tuning consistently outperforms PromptSUM across multiple STS benchmarks and downstream tasks. These results highlight the strong generalizability of our proposed method and its effectiveness in various application scenarios.

\section{Memory Usage and Batch Sizes}
\label{appendix:batch}

Here, we examine the memory consumption of Pcc-tuning and analyze the impact of batch size on model performance. For implementation, we utilized four 24GB NVIDIA GPUs, with PromptSUM as the manual template across all experimental groups.

Table~\ref{tab:memory} presents the memory usage for each of the two fine-tuning stages of Pcc-tuning across different backbone models. The results show that optimizing the model with Pearson's correlation coefficient in the second stage requires significantly fewer computational resources compared to contrastive learning in the first stage. Furthermore, given that the maximum sequence length supported in the second stage of Pcc-tuning is twice that of the contrastive learning stage, our method demonstrates a clear advantage in memory efficiency.
\begin{table}[htbp]
\centering
\begin{tabular}{c|c|c}
    \toprule
    \bf PLMs & \bf Stage  & \bf Memory (GB)\\
    \midrule
    \multirow{2}{*}{OPT$_{\rm 6.7b}$} 
    & I & 91.83 \\ 
    & II & 58.44 \\
    \midrule
    \multirow{2}{*}{LLaMA$_{\rm 7b}$} 
    & I & 93.82 \\
    & II & 65.22 \\
    \midrule
    \multirow{2}{*}{LLaMA2$_{\rm 7b}$}
    & I & 93.82 \\
    & II & 69.37 \\
    \midrule
    \multirow{2}{*}{Mistral$_{\rm 7b}$} 
    & I & 93.41 \\
    & II & 73.31 \\
    \bottomrule
\end{tabular}
\caption{Memory consumption for each stage of Pcc-tuning's two-stage training pipeline.}
\label{tab:memory}
\end{table}

Several factors contribute to this improvement, with batch size being a critical one. In standard InfoNCE Loss, negative instances for the current sample are drawn from other texts within the same batch. As a result, contrastive learning-based STS solutions typically require large batch sizes to provide sufficient reference information for optimizing embeddings. For example, SimCSE employs a batch size of 512 in supervised settings. Additionally, our experiments in section~\ref{sec:contrast}  also support this observation.

Given this context, we were interested in exploring how varying batch sizes affect performance when using Pearson's correlation coefficient for training. Therefore, we tested Pcc-tuning on seven STS tasks collected in SentEval, using four 7B-level generative PLMs under different batch size conditions. Due to the relatively slow inference speed of 7B-scale models, we did not perform an exhaustive grid search, but intuitively selected several batch sizes for testing, which was sufficient to illustrate the key trends.

The results are summarized in Table~\ref{tab:batch}. It is evident that even with variations in batch size by several dozen, Pcc-tuning maintains consistently high performance. This indicates that our proposed method is not sensitive to batch size. Combined with the findings from section~\ref{sec:prompts}, where Pcc-tuning exhibits minimal performance fluctuations under different prompts, we conclude that Pcc-tuning demonstrates exceptional robustness and can easily adapt to a wide range of hyperparameter configurations.
\begin{table}[htbp]
\centering
\begin{tabular}{c|c|c}
    \toprule
    \bf PLMs & \bf Batch Size  & \bf Spearman\\
    \midrule
    \multirow{4}{*}{OPT$_{\rm 6.7b}$} 
    & 192 & 86.88 \\ 
    & 200 & \bf 86.93 \\
    & 240 & 86.90 \\
    & 248 & 86.89 \\
    \midrule
    \multirow{4}{*}{LLaMA$_{\rm 7b}$} 
    & 192 & 87.34 \\
    & 200 & \bf 87.35 \\
    & 216 & 87.27 \\
    & 224 & 87.29 \\
    \midrule
    \multirow{4}{*}{LLaMA2$_{\rm 7b}$}
    & 200 & 87.77 \\
    & 208 & 87.73 \\
    & 216 & \bf 87.80 \\
    & 240 & 87.68 \\
    \midrule
    \multirow{4}{*}{Mistral$_{\rm 7b}$} 
    & 200 & 87.76 \\
    & 208 & \bf 87.86 \\
    & 216 & 87.75 \\
    & 256 & 87.73 \\
    \bottomrule
\end{tabular}
\caption{Pcc-tuning's average Spearman scores on seven STS benchmarks under different batch sizes.}
\label{tab:batch}
\end{table}

\end{document}